\documentclass{article}

\usepackage{fix-cm}
\usepackage{bm}
\usepackage{wrapfig}
\usepackage{multirow}
\usepackage{longtable}
\usepackage{supertabular}
\usepackage{colortbl}
\usepackage{subcaption}
\usepackage[most]{tcolorbox}
\usepackage{xcolor}
\definecolor{LightCyan}{rgb}{0.88,1,1}

\newcommand{\myPara}[1]{\vspace{5pt}\noindent$\bullet$~\textbf{#1} \ }
\newcommand{\supp}[1]{\textcolor{magenta}{#1}}
\def\model{MODA}

\definecolor{blue3}{HTML}{5D8DFD}
\definecolor{green3}{HTML}{88B06D}
\definecolor{orange3}{HTML}{F5A83D}
\definecolor{red3}{HTML}{F5433D}
\definecolor{blue1}{HTML}{AEC6FE}
\definecolor{green1}{HTML}{B3CDA2}
\definecolor{orange1}{HTML}{F9CB8A}
\definecolor{red1}{HTML}{FBA09D}
\definecolor{lightgray}{gray}{1.0}
\definecolor{cambriangray}{gray}{0.9}
\definecolor{darkgreen}{HTML}{4FAC5B}

\usepackage{microtype}
\usepackage{graphicx}
\usepackage{booktabs}
\usepackage{hyperref}

\usepackage[accepted]{icml2025}
\usepackage{amsmath}
\usepackage{amssymb}
\usepackage{mathtools}
\usepackage{amsthm}
\usepackage[capitalize,noabbrev]{cleveref}
\usepackage[textsize=tiny]{todonotes}

\icmltitlerunning{\model: MOdular Duplex Attention for Multimodal Perception, Cognition, and Emotion Understanding}

\begin{document}

\twocolumn[
\icmltitle{\model: MOdular Duplex Attention for\\Multimodal Perception, Cognition, and Emotion Understanding}
\icmlsetsymbol{ld}{\ddag}
\icmlsetsymbol{it}{\dag}

\begin{icmlauthorlist}
\icmlauthor{Zhicheng Zhang}{nku,nkusz,it}
\icmlauthor{Wuyou Xia}{nku}
\icmlauthor{Chenxi Zhao}{nku,it}
\icmlauthor{Yan Zhou}{ks}
\icmlauthor{Xiaoqiang Liu}{ks}
\icmlauthor{Yongjie Zhu}{ks,ld}
\icmlauthor{Wenyu Qin}{ks}
\icmlauthor{Pengfei Wan}{ks}
\icmlauthor{Di Zhang}{ks}
\icmlauthor{Jufeng Yang}{nku,nkusz}
\end{icmlauthorlist}

\icmlaffiliation{nku}{VCIP \& TMCC \& DISSec, College of Computer Science, Nankai University}
\icmlaffiliation{nkusz}{Pengcheng Laboratory}
\icmlaffiliation{ks}{Kuaishou Technology}

\icmlcorrespondingauthor{Jufeng Yang}{yangjufeng@nankai.edu.cn}
\vskip 0.3in
]
\printAffiliationsAndNotice{\icmlIntern \icmlLeader}

\begin{abstract}
Multimodal large language models (MLLMs) recently showed strong capacity in integrating data among multiple modalities, empowered by a generalizable attention architecture.
Advanced methods predominantly focus on language-centric tuning while less exploring multimodal tokens mixed through attention, posing challenges in high-level tasks that require fine-grained cognition and emotion understanding.
In this work, we identify the attention deficit disorder problem in multimodal learning, caused by inconsistent cross-modal attention and layer-by-layer decayed attention activation.
To address this, we propose a novel attention mechanism, termed MOdular Duplex Attention (\model), simultaneously conducting the inner-modal refinement and inter-modal interaction.
\model~employs a correct-after-align strategy to effectively decouple modality alignment from cross-layer token mixing.
In the alignment phase, tokens are mapped to duplex modality spaces based on the basis vectors, enabling the interaction between visual and language modality.
Further, the correctness of attention scores is ensured through adaptive masked attention, which enhances the model's flexibility by allowing customizable masking patterns for different modalities.
Extensive experiments on 21 benchmark datasets verify the effectiveness of \model~in perception, cognition, and emotion tasks.
\textbf{\textit{Source code and demo are available in \href{https://zzcheng.top/MODA}{\supp{https://zzcheng.top/MODA}}.}}
\end{abstract}

\section{Introduction}

Benefiting from the blossom of large language models~\cite{chiang2023vicuna,dubey2024llama,teknium2024hermes}, multimodal large language models (MLLMs) have shown strong capacity in integrating multimodal data as human~\cite{tong2024cambrian,brown2020language,bai2023qwen}, which illuminate a promising pathway toward Artificial General Intelligence (AGI).
Advanced effort has been devoted to constructing MLLM~\cite{achiam2023gpt}, focusing on exploring more insightful data curation, model tuning, and evaluation benchmarks.
As the controller of agent, MLLMs provide a natural solution by conducting content perception~\cite{liu2023llava}, understanding role cognition~\cite{dai2024mmrole}, and analyzing human emotion~\cite{yang2024emollm}.
One more step forward into AGI lies in high-level multimodal understanding like humans, including cognition and emotion.
Cognition, as a higher-level capability, requires the ability to model relationships and reasoning across modalities~\cite{dai2024mmrole,pessoa2022entangled}.
Beyond cognition, emotion understanding is another critical aspect of fine-grained multimodal comprehension~\cite{yang2024emollm,Zhang_2023_CVPR}.
These high-level multimodal tasks pose new challenges for MLLMs.

\begin{figure}[!tbp]
\centering
\vspace{-5pt}
\includegraphics[width=\linewidth]{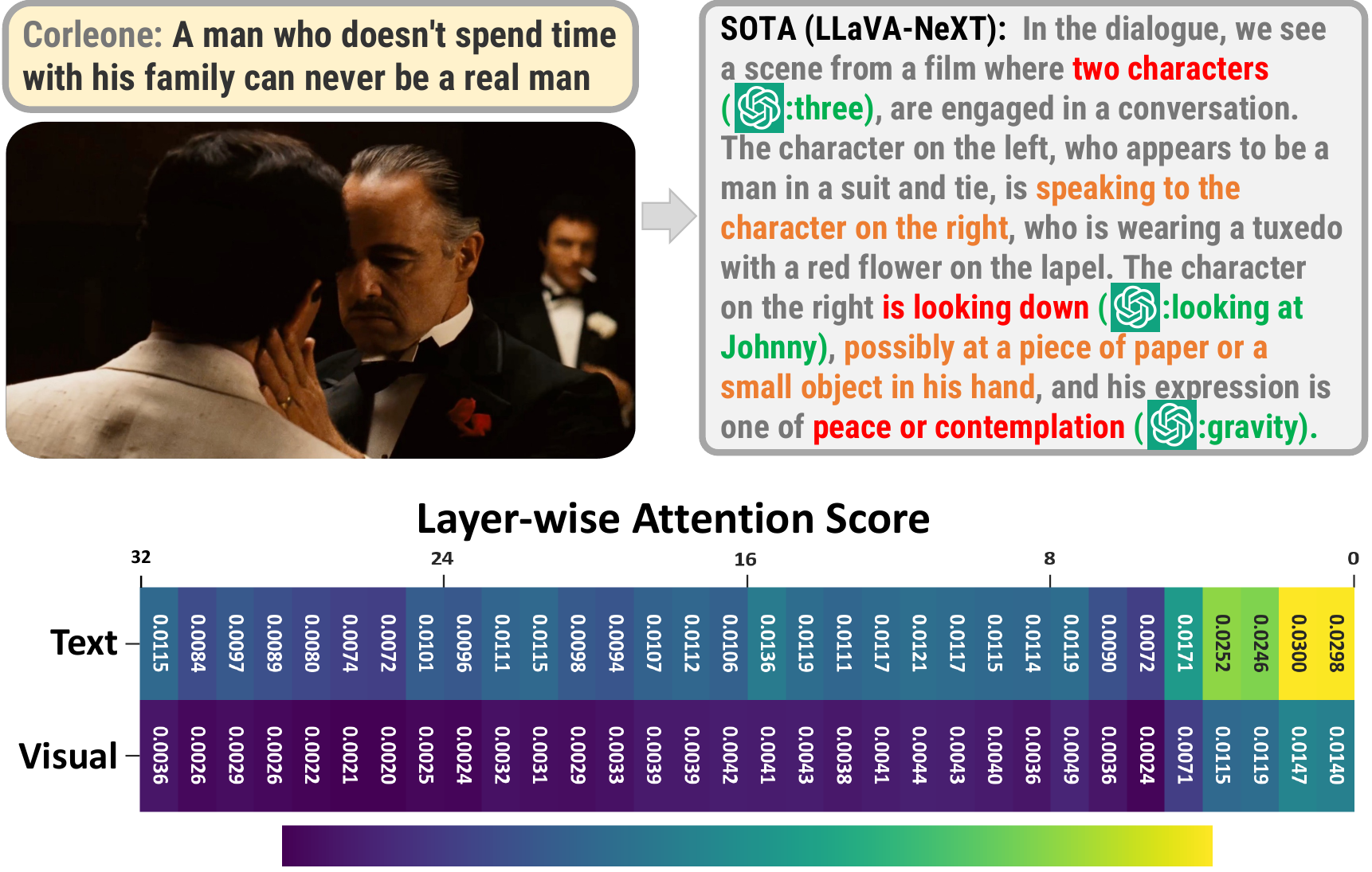}
  \put(-234,152){\small \textbf{(a)}}
  \put(-116,152){\small \textbf{(b})}
  \put(-234,58){\small \textbf{(c)}}
\vspace{-10pt}
\caption{\textbf{Illustration of deficit disorder attention problem.}
(a) Given the detailed image and lines from \textit{The Godfather}, (b) we highlight \textcolor{red}{incorrect responses}, corresponding \textcolor{orange}{hallucinated explanations}, and attached \textcolor{darkgreen}{answers}.
(c) We visualize attention score across layers, highlighting inconsistent attention across modalities.
}
\vspace{-5pt}
\label{fig:Motivation}
\end{figure}

While recent MLLMs show promising results in basic perception, they still struggle to perceive fine-grained details~\cite{tong2024eyes}, which is essential for understanding cognition and emotion.
Public benchmarks reveal that these advanced  MLLMs can underperform relative to random guessing~\cite{yang2024emollm}, with 3 SOTAs achieving approximately 50:50 accuracy in 2-class sarcasm detection on the HFM dataset.
This phenomenon arises from an excessive emphasis on the dominant modality data, leading to neglect of fine-grained details in alternative modality.

We delve deep into the reason and analyze the multimodal tokens mixed by attention in MLLM.
As shown in Fig.~\ref{fig:Motivation} (a)\&(b), we observe that SOTA MLLM struggles to capture fine-grained details (\textit{e.g.}, eyesights of character), leading to error in emotion understanding.
The reason behind this is inconsistent attention across multiple layers in MLLM (63\% disparity in Fig.~\ref{fig:Motivation} (c)), which we call deficit disorder attention problem.
On the one hand, the attention scores in MLLM exhibit a bias towards the language component.
On the other hand, layer-by-layer decay of attention further accentuates this disparity.
As a result, the attention score disparity across modalities can reach up to 10 times.

Our intuition is that multimodal attention mechanisms often suffer from imbalances between self-modal and cross-modal interactions, leading to suboptimal feature co-operation across modalities.
By explicitly separating and modulating these two components, we can better align multimodal features while preserving the unique characteristics of each modality.
To achieve this, we propose MOdular Duplex Attention (MODA), which splits attention into self-modal and cross-modal parts, each with its own modulated attention mask.
The self-modal attention component focuses on capturing the intrinsic relationships within individual modalities.
In contrast, the cross-modal attention component is responsible for aligning features across different modalities, facilitating effective information exchange.
At the core of the MODA model is the Duplex (V/T)-Aligner, which maps the tokens into a shared dual-modality representation space defined by two gram matrices.
Additionally, the Modular Masked Attention component allows the model to adaptively focus on relevant modalities by applying customized masking patterns, further enhancing its flexibility on multimodal understanding tasks.

Our contributions are two-fold as follows:
(1) From a novel perspective of the attention shift mechanism, we indicate the key bottleneck of attention among SOTA MLLMs and analyze the core reason in depth. We further propose a  modular and duplex attention mechanism based on our observation.
(2) We investigate a new MLLM for perception, cognition, and emotion, enabling applications in fine-grained understanding and planning. Extensive experiments on 21 benchmarks verify the generalization and effectiveness of~\model.

\section{Related Work}

\noindent\textbf{Multimodal large language model (MLLM)}
have garnered significant attention recently due to their ability to integrate pre-trained foundational models, especially powerful Large Language Models (LLMs)\cite{achiam2023gpt,touvron2023llama}, alongside multimodal encoders\cite{dosovitskiy2021an,radford2021learning}.
These models enhance the processing of multimodal inputs and outputs, as demonstrated in advanced works~\cite{alayrac2022flamingo,bai2023qwen}.
MLLMs leverage attention mechanisms to facilitate multimodal token mixing, enabling both inductive and deductive understanding across modalities.
However, the vision modality's potential remains underutilized in many of these models.
MMVP~\cite{tong2024eyes} identifies a critical issue, highlighting how existing MLLMs fail to fully activate the vision modality due to improper handling of low-level visual attributes.
Further, Cambrian-1~\cite{tong2024cambrian} confirms this limitation and introduces a spatial vision aggregator to enhance visual feature.
In this work, we investigate the root cause of these limitations, identifying the bottleneck in the design of the multimodal attention mechanism.
To address the issue of imbalanced attention scores, we propose a novel multimodal attention that better balances the contributions of each modality.

\noindent\textbf{Understanding cognition and emotion}
~\cite{fu2023mme,yang2024emollm} play an important role in the pathway toward building an intelligent agent, except for content understanding demonstrated by prior MLLMs.
As two of high-level understanding,
cognition~\cite{wang-etal-2024-incharacter,kong2023better,salemi-etal-2024-lamp} typically refers to the ability to make decisions and judgments similar to characters~\cite{binz2023using,wang2023rolellm,deshpande2023toxicity}, such as generating website code~\cite{zhu2023minigpt,wang2024mllmtool}, or role playing~\cite{chen2024persona,zhang-etal-2018-personalizing}.
Emotion mainly depends on the psychology assumptions~\cite{zhao2021emotion,zhang2024masked}, where the categorical one is mostly used due to it being easily understandable~\cite{yang2018weakly,mai2022hybrid,lian2022smin,10.1145/3503161.3548007}.
However, it is less explored due to its requirements for fine-grained content understanding, which MLLMs can hardly achieve.

\begin{figure*}[!tbp]
\centering
\includegraphics[width=.99\linewidth]{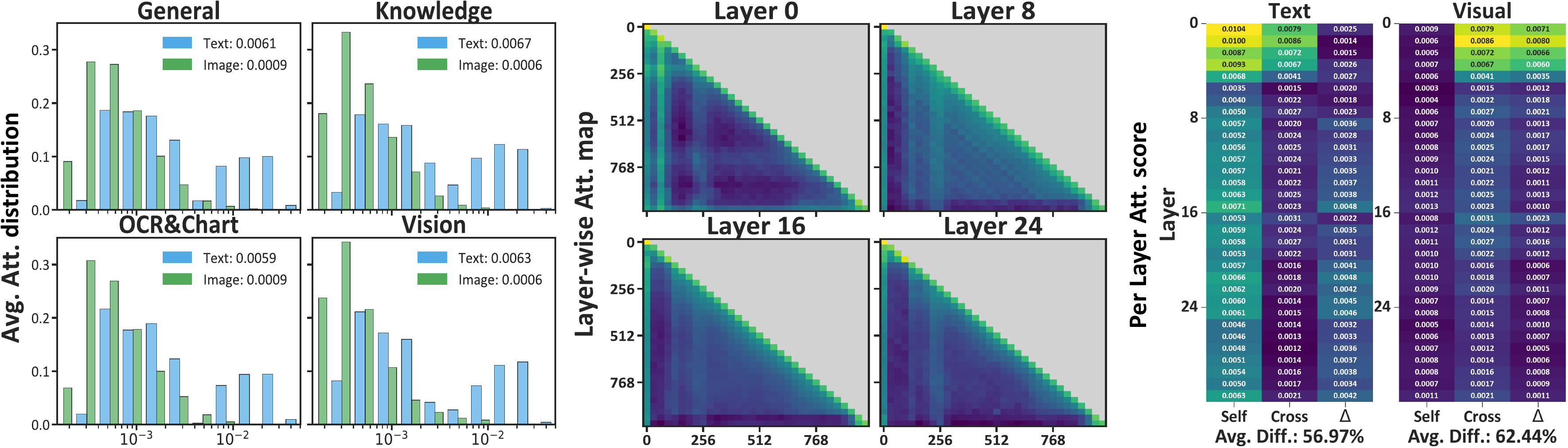}
\put(-467,136){\small \textbf{(a)}}
\put(-287,136){\small \textbf{(b)}}
\put(-115,136){\small \textbf{(c)}}
\caption{Analysis of existing MLLMs on four fine-grained understanding tasks. (a) The distribution of attention activation values among visual and textual tokens. (b) The attention map for multimodal tokens among stages. (c) The self- and cross-modal attention activation scores with their disparity among the attention layers.}
\label{fig:Analysis}
\vspace{4pt}
\end{figure*}
\noindent\textbf{Attention in MLLM}
plays a critical role in addressing the computational and memory challenges inherent in their design. 
Significant progress has been made in developing efficient attention mechanisms for Transformer architectures, which include fixed patterns~\cite{child2019generating}, combinations of patterns~\cite{zaheer2020big}, learnable patterns~\cite{kitaev2020reformer}, neural memory~\cite{beltagy2020longformer}, low-rank methods~\cite{wang2020linformer}, and kernel-based techniques~\cite{choromanski2020rethinking}.
For example, the Set Transformer introduces inducing points to handle set-input problems~\cite{wang2020linformer}, while the Axial Transformer applies attention along individual axes of input tensors, reducing computational overhead~\cite{beltagy2020longformer}.
These innovations collectively enhance the scalability of Transformer models, enabling their application to tasks with large inputs or long sequences~\cite{choromanski2020rethinking,han2024agent}.
While previous approaches have focused on improving the efficiency and scalability of attention in single-modal tasks, the multimodal context introduces unique challenges, such as balancing attention scores across heterogeneous modalities~\cite{zhao2021emotion}.
Our work extends this line of research by specifically addressing the multimodal attention mechanism in MLLMs.

\section{Methodology}
\label{sec:form}

\subsection{Preliminary}
\myPara{Attention}
Given the input multimodal tokens, $\bm{X} \in \mathbb{R}^{N \times d}$, $N$ be the number of tokens and $d$ be the dimensionality of the hidden state.
Let $\bm{A} \in \mathbb{R}^{N \times N}$ denote the attention score matrix computed among $N$ tokens, we have $\bm{A}={\bm{Q}\bm{K}^{\top}} / {\tau}$, and the output of attention layer as:
\begin{align}
\bm{O} &= \text{Softmax} ( \frac{\bm{Q}\bm{K}^{\top}}{\tau} + \bm{M} ) \bm{V}.
\end{align}
where $\bm{Q},\bm{K},\bm{V}\in\mathbb{R}^{d \times d}$ represents query, key, and value matrix derived from input tokens.
Attention is also practically masked $\bm{M}\in\mathbb{R}^{N \times N}$ to filter out special tokens~\cite{li2023blip} or conduct causal sequential modeling~\cite{wang2024emu3,achiam2023gpt}.

\myPara{Multimodal Attention}
Formally, consider a multimodal token sequence $\bm{X}_M$ comprising $M$ modalities.
The total token length is $N_M=N_1+\dots+N_M$, where $N_m$ represents the length of the $m^{th}$ modality token sequence $\bm{X}_m$.
The attention can be split into two parts for each modality token sequence, self-modal attention and cross-modal attention.
We have ${(\cdot)}^{[m,\bar{m}]}$, which represents the tokens derived from the $m^{th}$ modality and rest.
For the self-modal and cross-modal attention, we have
\begin{align}
\bm{O}_{self} = \text{Softmax} ( \frac{\bm{Q}^{m}{\bm{K}^{m}}^{\top}}{\tau} + \bm{M} ) \bm{V}^{m},\\
\bm{O}_{cross} = \text{Softmax} ( \frac{\bm{Q}^{m} {\bm{K}^{\bar{m}}}^{\top}}{\tau} + \bm{M} ) \bm{V}^{\bar{m}}.
\end{align}

\subsection{Deficit Disorder Attention Problem}
\label{sec:Analysis}
Recently, multimodal attention has played a very important role in multimodal areas, including diffusion models that involve cross-modal generation and MLLM that involves cross-modal understanding.
The attention mechanism governs token interactions by computing similarities and applying masks.
To further investigate the Attention Deficit Disorder (DDA) phenomenon, we conduct a series of analyses on four categories of fine-grained understanding tasks.

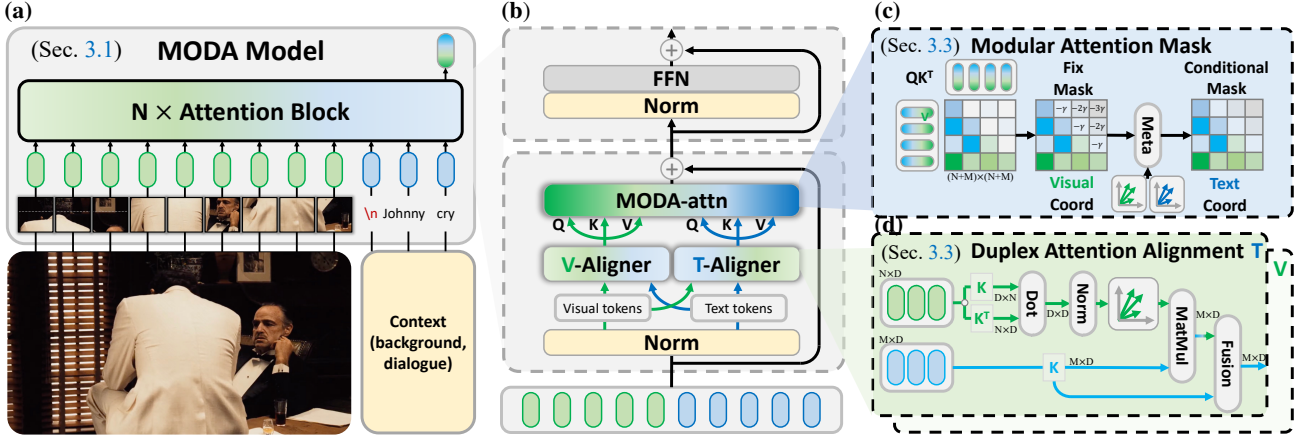
\begin{figure*}[!tbp]
  \centering
  \includegraphics[width=\linewidth]{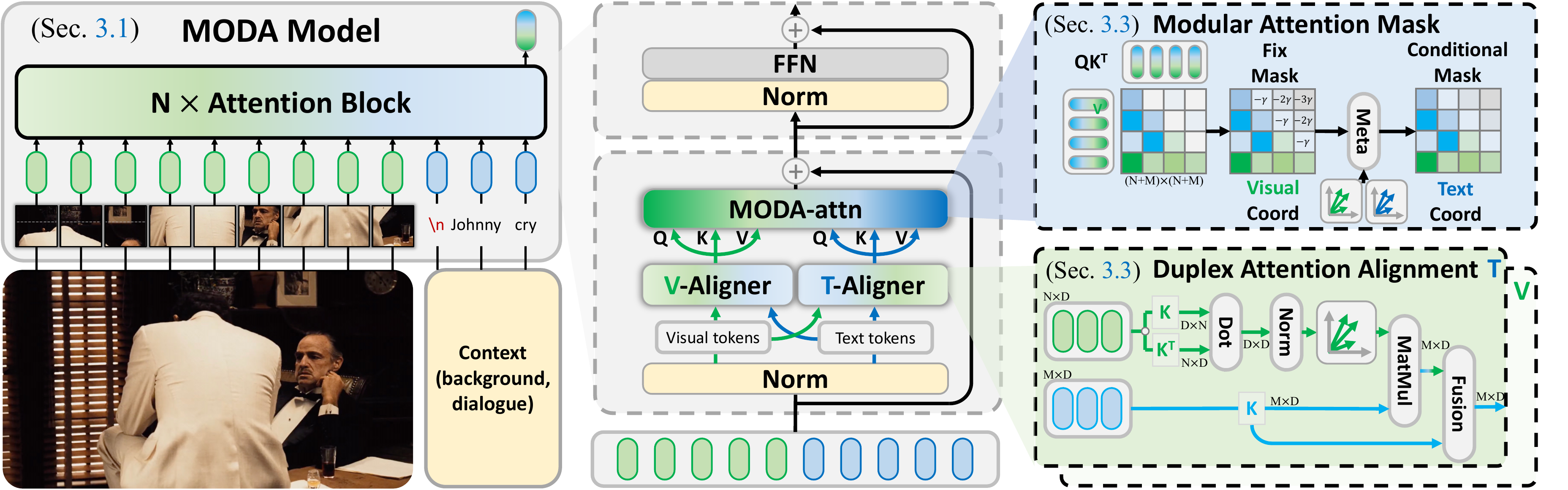}
  \put(-488,158){\small \textbf{(a)}}
  \put(-300,158){\small \textbf{(b})}
  \put(-159,158){\small \textbf{(c)}}
  \put(-159,77){\small \textbf{(d)}}
  \caption{MOdular Duplex Attention. (a) \model~takes the image and contextual prompt as input, including the background and history of the conversation. (b) With~\model, the token flows are justified in each Transformer block of MLLM. \model~ modifies the deficient attention scores in a correct-after-align manner via (c) Modular masked attention and (d) Duplex (V/T)-aligner.}
  \label{fig:pipeline}
\end{figure*}
As shown in Fig.~\ref{fig:Analysis} (a), we observe that the attention devoted to visual content is significantly weaker compared to that for the textual modality.
This observation aligns with the challenges faced by MLLMs fine-tuned from autoregressive models in handling fine-grained visual perception.
The inherent design of MLLM, which is primarily optimized for text-based tasks, may lead to an underrepresentation of visual features when extended to multimodal contexts.
This imbalance highlights a critical limitation in the current architecture, where the model's proficiency in textual processing does not seamlessly translate to an equivalent capability.
Further, we conduct experiments on Fig.~\ref{fig:Analysis} (b)\&(c), and we observe a distinct cross-attention bias in the lower layers of the model across its 32 layers.
This bias is notably inconsistent with the distribution of attention in the higher layers, which are known for their stronger representational capabilities.
Specifically, the lower layers tend to focus disproportionately on cross-modal interactions, potentially at the expense of effectively capturing intra-modal features, leading to suboptimal multimodal integration.

This leads to the formal introduction of the Deficit Disorder Attention (DDA) problem.
Given the visual tokens $x_v^l$ and text tokens $x_t^l$ in the block $l$, the multimodal attention builds the link from two parts (\textit{i.e.}, self-modal $x_t^l \rightarrow x_t^{l+1}, x_v^l \rightarrow x_v^{l+1}$ and cross-modal $x_t^l \rightarrow x_v^{l+1}, x_v^l \rightarrow x_t^{l+1}$ ),  where the links are commonly implemented by the pair-wise token similarity and weighted sum.
However, the modality gap between tokens decrease the magnitude of links, as we observed, the link value of $x_v^l \rightarrow x_v^{l+1}$ and $x_t^l \rightarrow x_v^{l+1}$ decays exponentially with depth ($\alpha_{v,t \rightarrow v}^l \propto \gamma^l,\gamma \neq 1$).
This misalignment propagates layer-wise, causing the cumulative error in cross-modal interaction to grow as 
\begin{equation}
\mathbb{E}_{DDA} = \prod_l \gamma^l \epsilon_l,
\end{equation}
where $\epsilon_l$ denotes the layer-specific alignment error.
This phenomenon aligns with the theoretical insights in~\cite{dong2021attention}, where pure attention mechanisms experience rank collapse, a critical factor that exacerbates the imbalance in attention distribution.

\subsection{MOdular Duplex Attention}
When the gap across modalities arises, we propose to align the tokens from multiple modalities in the attention, which we call modular duplex attention (\model).
\model~first splits multimodal attention into the modality alignment part and the token focus correction part.

\myPara{Duplex Attention Alignment}
To reduce the modality inconsistency, a natural idea is to align them.
Inspired by the recent advance of visual-language embedding space mapping in diffusion model~\cite{rombach2022high}, we propose mapping the token into the other modality space according to the embedding space bases of the gram matrix.
We extract the basis vector of each modality space according to the gram matrix of tokens~\cite{ryu2023gramian,peebles2023scalable}, thus compressing the semantics of each modality and serving as a transfer for other modalities.
Thus, the duplex attention alignment consists of V-Aligner and T-Aligner responsible for visual and language modality, respectively.
Specifically, for the $m^{th}$ modality, the space bases are given according to the normed gram matrix $||\bm{G}^m||\in\mathbb{R}^{d\times d}$, where $\bm{G}^m_{ij}$ is the inner product between tokens $i$ and $j$:
\begin{align}
    \bm{G}^m_{ij} = \sum_{k=1}^{N_m}{\bm{K}^m_{ik}\bm{K}^m_{kj}}= {\bm{K}^m}^\top \bm{K}^m,
\end{align}
where $\bm{K}^m$ are the key states of the $m^{th}$ modality tokens and $N_m$ is the number of token belong to modality $m$.
By including the base vectors of the space defined by the Gram matrix, we can effectively capture the relationships among the tokens within the $m^{th}$ modality.
This allows us to construct a feature representation that is not only rich in information but also maintains the intrinsic structure of the data. 
\begin{table*}[!tbp]
    \centering
    \captionsetup[subfloat]{labelformat=empty}
    \centering
    \caption{\textbf{Ablation Study.}
    We conduct experiments on four types of multimodal tasks, including general QA (\textbf{G}), knowledge QA (\textbf{K}), OCR\&Chart QA (\textbf{O}), and vision-centric QA (\textbf{V}).
    The lines with blue shallow indicate the optimal setting for our method.
    If not otherwise specified, this setting is used for all subsequent experiments.
    }
    \vspace{2pt}
    \begin{minipage}[c]{\linewidth}
    \hspace{-8pt}
        \subfloat[(a) \textbf{Module}]
        {
        \begin{minipage}[c]{.293\linewidth}
        \centering
        
        \resizebox{!}{3.em}{
        \begin{tabular}{
        p{0.8cm}<{\centering}
        p{0.8cm}<{\centering}
        |
        p{.55cm}<{\centering}
        p{.55cm}<{\centering}
        p{.55cm}<{\centering}
        p{.55cm}<{\centering}
        }
        \toprule
        MDM&
        DAA&
        \textbf{G}&
        \textbf{K}&
        \textbf{O}&
        \textbf{V}
        \\
        \midrule
        
        -&-
        &63.6&44.0&60.8&38.0\\
        
        \checkmark&-
        &69.2&45.4&60.9&42.6\\
        
        -&\checkmark
        &67.8&47.6&63.3&48.1\\
        
        \rowcolor{LightCyan}
        \checkmark&\checkmark
        &69.3&48.3&67.0&54.3\\
        \bottomrule
        \end{tabular}
        }
        \end{minipage}
        }
        \subfloat[(b) \textbf{Attention Alignment}]
        {
        \begin{minipage}[c]{.245\linewidth}
        \centering
        
        \resizebox{!}{3.em}{
        \begin{tabular}{
        p{1.3cm}<{\centering}
        |
        p{.45cm}<{\centering}
        p{.45cm}<{\centering}
        p{.45cm}<{\centering}
        p{.45cm}<{\centering}
        }
        \toprule
        align&
        \textbf{G}&
        \textbf{K}&
        \textbf{O}&
        \textbf{V}
        \\
        \midrule

        \multicolumn{1}{l|}{MLP}
        &69.5&47.5&66.8&46.0\\
        
        \multicolumn{1}{l|}{\ +2xMLP}
        &66.5&48.6&67.9&49.1\\

        \multicolumn{1}{l|}{\ +GeLU}
        &69.5&49.1&64.0&54.2\\
        
        \rowcolor{LightCyan}
        \multicolumn{1}{l|}{\ +CoV}
        &69.3&48.3&67.0&54.3\\
        
        \bottomrule
        \end{tabular}
        }
        \end{minipage}
        }
        \subfloat[(c) \textbf{Attention Fusion}]
        {
        \begin{minipage}[c]{.22\linewidth}
        \centering
        
        \resizebox{!}{3.em}{
        \begin{tabular}{
        p{0.75cm}<{\centering}
        |
        p{.45cm}<{\centering}
        p{.45cm}<{\centering}
        p{.45cm}<{\centering}
        p{.45cm}<{\centering}
        }
        \toprule
        fusion&
        \textbf{G}&
        \textbf{K}&
        \textbf{O}&
        \textbf{V}
        \\
        \midrule
        $X_p$
        &69.2&45.4&60.9&42.6\\

        $X_a$
        &67.8&47.6&63.3&48.1 \\

        \rowcolor{LightCyan}
        Con
        &69.3&48.3&67.0&54.3\\
        
        Add
        &62.2&47.6&67.2&52.2 \\

        \bottomrule
        \end{tabular}
        }
        \end{minipage}
        }
        \subfloat[(d) \textbf{Attention Mask}]
        {
        \begin{minipage}[c]{.21\linewidth}
        \centering
        
        \resizebox{!}{3.em}{
        \begin{tabular}{
        p{0.6cm}<{\centering}
        |
        p{.45cm}<{\centering}
        p{.45cm}<{\centering}
        p{.45cm}<{\centering}
        p{.45cm}<{\centering}
        }
        \toprule
        mask&
        \textbf{G}&
        \textbf{K}&
        \textbf{O}&
        \textbf{V}
        \\
        \midrule

        Inf
        &67.8&47.6&63.3&48.1\\

        Fix
        &70.1&49.0&67.0&52.3 \\
        
        \rowcolor{LightCyan}
        Attn.
        &69.3&48.3&67.0&54.3\\
        
        \texttt{[M]}
        &69.5&47.5&66.8&46.0\\

        \bottomrule
        \end{tabular}
        }
        \end{minipage}
        }

    \end{minipage}
    \label{tab:ablation}
\end{table*}

As a following product, the normed gram matrix serves as a cross-modal token transfer function, enabling an efficient transformation of tokens from other modality $\bar{m}$ into the modality $m$ as a kernelized mapping function $f:\mathbb{R}^{d}\rightarrow\mathbb{R}^{d}$.
The aligned tokens are computed as follows:
\begin{align}
    \bm{K}^{\bar{m} \to m} = \bm{K}^{\bar{m}} ||\bm{G}^m|| ,
\end{align}
where $\bm{K}^{\bar{m}}$ represents the value representation from other modalities $\bar{m}$.
The mapped tokens are further fused with the original ones to enhance the token similarity among all the modalities.
Due to the substantial computational expense associated with training a complete MLLM, we utilize token merging and LoRA-based tuning to develop the fuser.
Notably, the computation in the alignment stage keeps linear complexity to the token number, since the matrix sum among tokens is only conducted in the first round.

\myPara{Modular Attention Mask}
Attention mask controls the flow of tokens across transformer layers and induces the positional bias for MLLM~\cite{wu2024on}.
To better fit the requirements of the multimodal token sequence,
we assign a modulated attention mask for each modality, where the mask is split into  $\bm{M}^{m}$ and $\bm{M}^{\bar{m}}$ responsible for self- and cross-modality, respectively.
\begin{align}
\bm{O}_{self} = \text{Softmax} ( \frac{\bm{Q}^{m}{\bm{K}^{m}}^{\top}}{\tau} + \bm{M}^{m} ) \bm{V}^{m},\\
\bm{O}_{cross} = \text{Softmax} ( \frac{\bm{Q}^{m} {\bm{K}^{\bar{m}}}^{\top}}{\tau} + \bm{M}^{\bar{m}} ) \bm{V}^{\bar{m}}.
\end{align}

To alleviate the collapsed attention matrix and prevent it from under-smoothed tokens.
We first introduce a modular attention mask that stores unnecessary attention values as pseudo-attention scores~\cite{10.5555/3692070.3694424}.
For each row, representing the attention scores for the $i$-th token, the sequence length that the token can attend to is fixed at $n$.
Consequently, each row contains $n-i$ pseudo-attention scores, which are allocated to the excess values.
The attention scores are formally represented using a masking strategy with a decay rate $\beta$, as follows:
\begin{align}
A_{MM} &=
\begin{pmatrix}
\bm{q}_{1}\bm{k}_{1}^{\top} & p_{11} & \cdots & p_{1(n-1)} \\
\bm{q}_{2}\bm{k}_{1}^{\top} & \bm{q}_{2}\bm{k}_{2}^{\top} & \cdots & p_{1(n-2)} \\
\vdots & \vdots & \ddots & \vdots \\
\bm{q}_{n}\bm{k}_{1}^{\top} & \bm{q}_{n}\bm{k}_{2}^{\top} & \cdots & \bm{q}_{n}\bm{k}_{n}^{\top}
\end{pmatrix}\\
p_{base}&=0, p_{ij} = p_{base} - (j - 1)\beta
\end{align}

Except for the absolute location prior information, we further introduce the modality location to enforce the model to correct the token flow.
We introduce the normed gram matrix as an indicator, to find out the part should be carried with modality location priors.
We introduce the normed Gram matrix to serve as a critical indicator, guiding the model in identifying which components should leverage modality location priors.
This separation allows for more precise control over how tokens from the same modality interact with each other versus how they engage with tokens from other modalities.
The self-modal attention, represented by $\bm{O}_{self}$, focuses on refining the relationships within the same modality, ensuring that relevant information is effectively propagated through the layers. Conversely, the cross-modal attention, denoted by $\bm{O}_{cross}$, facilitates the exchange of information between distinct modalities, enabling the model to leverage complementary features.

\section{Experiment}

\subsection{Benchmark Datasets}

\textbf{\textit{Perception}}:
Following~\cite{tong2024cambrian}, we conduct experiments on 4 types of perception task (\textit{i.e.}, general, knowledge, ocr, and vision-centric) across 16 benchmarks: MME~\cite{fu2023mme}, MMBench~\cite{liu2025mmbench}, SEED~\cite{li2023seed}, GQA~\cite{hudson2019gqa}, ScienceQA~\cite{lu2022learn}, MMMU~\cite{yue2024mmmu}, MathVista~\cite{lu2023mathvista}, AI2D~\cite{kembhavi2016diagram}, ChartQA~\cite{masry2022chartqa}, OCRBench~\cite{liu2023hidden}, TextVQA~\cite{singh2019towards}, DocVQA~\cite{mathew2021docvqa}, MMVP~\cite{tong2024eyes}, RealworldQA~\cite{grok}, and CV-Bench~\cite{tong2024cambrian}.
We adopt GPT4 score to evaluate response.

\noindent \textbf{\textit{Cognition}}: 
Following~\cite{dai2024mmrole}, we conduct experiments on MMRole to evaluate role-playing performance from 8 aspects: instruction adherence, fluency, coherency, image-text relevance, response accuracy, personality consistency, knowledge consistency, and tone consistency.

\noindent \textbf{\textit{Emotion}}: 
Following~\cite{10096777,10.1145/3652583.3658115}, we conduct experiments on 4 benchmark datasets.
MVSA-S and MVSA-M~\cite{MVSA} are datasets used for sentiment polarity classification (positive or negative), while TumEmo~\cite{9246699} is a multimodal dataset designed for classifying six basic emotions.
Additionally, HFM~\cite{HFM} is a multimodal dataset focused on recognizing high-level implicit emotion of sarcasm.
\begin{table*}[!tbp]
\centering
\vspace{8pt}
\fontsize{4pt}{4.8pt}\selectfont
\setlength\tabcolsep{2pt}
\renewcommand{\arraystretch}{1.1} 
\scalebox{1.76}{

\begin{tabular}{r |ccccc |ccccc |ccccc |ccccc}
\multicolumn{1}{c}{Model} &
\multicolumn{5}{c}{General} &
\multicolumn{5}{c}{Knowledge} &
\multicolumn{5}{c}{OCR \& Chart} &
\multicolumn{5}{c}{Vision-Centric}  \\
Method &
\rotatebox{90}{Avg} &
\rotatebox{90}{MME$^\text{P}$} &
\rotatebox{90}{MMB} &
\rotatebox{90}{SEED$^\text{I}$} &
\rotatebox{90}{GQA} &
\rotatebox{90}{Avg} &
\rotatebox{90}{SQA$^\text{I}$} &
\rotatebox{90}{MMMU$^\text{V}$} &
\rotatebox{90}{MathVista$^\text{M}$} & \rotatebox{90}{AI2D} &
\rotatebox{90}{Avg} &
\rotatebox{90}{ChartQA} &
\rotatebox{90}{OCRBench} &
\rotatebox{90}{TextVQA}  &
\rotatebox{90}{DocVQA}  &
\rotatebox{90}{Avg} &
\rotatebox{90}{MMVP} &
\rotatebox{90}{RealworldQA} &
\rotatebox{90}{CV-Bench$^\text{2D}$} &
\rotatebox{90}{CV-Bench$^\text{3D}$} \\
\hline

GPT-4V  & 63.0 & 1409.4 & 75.8 & 69.1 & 36.8 & 65.2 & 75.7 & 56.8 & 49.9 & 78.2 & 77.4 & 78.5 & 64.5 & 78.0 & 88.4 & 62.4 & 50.0 & 61.4  & 64.3 & 73.8   \\

Gemini-1.0 Pro & - & 1496.6 & 73.6 & 70.7 & - & - &79.5& 47.9 & 45.2 & -& - & - & 65.9 & - & - &- & - & -  & - &-\\
Gemini-1.5 Pro & - & - &- & - & - & - & - & 58.5 & 52.1 & 80.3 & - & 81.3 &- & 73.5 & 86.5 & - & - & 67.5 & - & -  \\
Grok-1.5 & - & - &- & - & - & - & - & 53.6 & 52.8 & 88.3 & - & 76.1 &- & 78.1 & 85.6 & - & - & 68.7 & - & -  \\

MM-1-8B & - & 1529.3 & 72.3 & 69.9 & - & - & 72.6 & 37.0 & 35.9 & - & - & - & - & - & - & - & - & - & - & -  \\
MM-1-30B & - & 1637.6 & 75.1 & 72.1 & - & - & 81.0 & 44.7 & 39.4 & - & - &- & - & - &- & - &- & - & - & -   \\
\hline
\rowcolor{gray!10}
\multicolumn{1}{l|}{\textit{Base LLM: Llama-3-Ins-8B}} &  &  &  &  &  &  &  &  &  &  &  &  &  &  &  &  &  &  &  &  \\
Mini-Gemini-HD-8B  & {72.7} & \textbf{1606.0} & 72.7 & 73.2 & 64.5 & 55.7 & 75.1 & 37.3 & 37.0 & {73.5} & 62.9 & 59.1 & 47.7 & 70.2  & 74.6 & 51.5 & 18.7 & 62.1 & 62.2 & 63.0  \\

LLaVA-NeXT-8B & 72.5 & {1603.7} & 72.1 & 72.7 & \textbf{65.2} & 55.6 & 72.8 & 41.7 & 36.3 & 71.6 & 63.9 & 69.5 & 49.0 & 64.6  & 72.6 & 56.6 & 38.7 &  60.1 & 62.2 & 65.3 \\

Cambrian-1-8B  & \textbf{73.1} & 1547.1 & \textbf{75.9} & {74.7} & {64.6} & {61.3} & \textbf{80.4} & {42.7} & \textbf{49.0} & {73.0} & {71.3} & {73.3} & {62.4} & \textbf{71.7} & {77.8} & {65.0} & {51.3} & \textbf{64.2} & {72.3} & {72.0} \\

\rowcolor{LightCyan}
\model-8B  & 72.1 & 1535.9 & {73.8} & \textbf{74.9} & 63.0 & \textbf{61.5} & \textbf{80.4} & \textbf{43.1} & {48.8} & \textbf{73.6} & \textbf{72.0} & \textbf{74.3} & \textbf{65.2} & {70.4} & \textbf{78.1} & \textbf{66.0} & \textbf{52.6} & {64.1} & \textbf{73.5} & \textbf{73.8} \\

\rowcolor{gray!10}
\multicolumn{1}{l|}{\textit{Base LLM: Hermes2-Yi-34B}} &  &  &  &  &  &  &  &  &  &  &  &  &  &  &  &  &  &  &  &  \\

Mini-Gemini-HD-34B & 76.2 & {1659.0} & 80.6 & 75.3 & 65.8 & 62.4 & 77.7 & 48.0 & 43.4 & {80.5} & 68.1 & 67.6 & 51.8 & 74.1 & \textbf{78.9} & 63.8 & 37.3 & 67.2 & 71.5 & 79.2 \\

LLaVA-NeXT-34B & 76.0 & 1633.2 & 79.3 & \textbf{75.9} & \textbf{67.1} & 62.5 & 81.8 & 46.7 & 46.5 & 74.9 & 67.7 & 68.7 & 54.5 & 69.5 & 78.1 & 64.0 & 47.3 & 61.0 & 73.0 & 74.8 \\

Cambrian-1-34B  & \textbf{76.8} & \textbf{1689.3} & {81.4} & 75.3 & 65.8 & {67.0} & {85.6} & {49.7} & {53.2} & 79.7 & {71.9} & {75.6} & {60.0} & {76.7} &  75.5 & {68.5} & {52.7} & {67.8} & {74.0} & {79.7} \\

\rowcolor{LightCyan}
\model-34B  & 76.7 & 1639.2 & \textbf{82.3} & {75.8} & {66.2} & \textbf{69.5} & \textbf{88.1} & \textbf{52.5} & \textbf{54.0} & \textbf{83.4} & \textbf{74.7} & \textbf{79.8} & \textbf{62.7} & \textbf{78.3} & {78.2} & \textbf{69.9} & \textbf{53.8} & \textbf{68.5} & \textbf{75.8} & \textbf{81.3} \\

\end{tabular}
}
\caption{\textbf{Comparison of \model\ with other leading MLLM framework on twelve perception benchmarks.} \model\  outperforms other open-source models and achieves competitive performance on a number of benchmarks, compared to proprietary models such as GPT-4V, Gemini, and Grok-1.5. The reported numbers of leading MLLMs come from ~\cite{tong2024cambrian}.}
\label{tab:comp_perception}
\end{table*}

\subsection{Settings}
We set the same experiment setting as~\cite{tong2024cambrian,liu2023llava}.
We adopt CLIP (ViT-L/14)~\cite{radford2021learning} as the visual encoder.
For the foundational large language model, we choose models from different scales, \textit{i.e.}, 8B: Llama-3-Instruct-8B~\cite{dubey2024llama} and 34B: Hermes2-Yi-34B~\cite{young2024yi}.
\model\ is trained for 1 epoch with a batch size of 2048, using the AdamW~\cite{loshchilov2018decoupled} optimizer with a cosine
learning rate schedule.
The learning rate is set to 2e-5 for LLM and 2e-6 for visual encoder, respectively.
The warmup rate is 0.03.

\subsection{Ablation Study}
To investigate the effectiveness of duplex attention alignment and modular attention mask, we conduct a component-wise ablation study in~\cref{tab:ablation}.
For ablation studies, we train the MLLMs at the scale of 8B, with the base LLM of Llama-3-Ins-8B.
For a fair comparison, all models are trained on 700K data samples for 1 epoch.
We further discuss each component by conducting in-depth analyses of their variants to answer the following research questions.
\begin{itemize}
\item \textbf{RQ1:} How does the design of duplex attention alignment impact cross-modal feature transfer?
\item \textbf{RQ2:} How does the modular attention mask address modality position bias and improve attention?
\item \textbf{RQ3:} How do the proposed duplex attention alignment and modular attention mask respectively interact to enhance multimodal attention?
\end{itemize}

\myPara{Response to RQ1: Modality Axis Transfer}
we analyze the effectiveness of duplex attention alignment in facilitating cross-modal feature transfer by examining its ability to align modality-specific features along a shared latent axis.
This is motivated by the need to reduce modality gaps and ensure effective information exchange between modalities.
We design experiments to test different variants of duplex attention alignment, such as using covariance matrices, attention head configurations, and linear vs. non-linear transformations.

\myPara{Response to RQ2: Modality Position Bias}
we investigate the role of the modular attention mask in addressing modality position bias and improving attention distribution.
This analysis is crucial for understanding how the mask prevents attention collapse and ensures balanced contributions from all modalities.
We experiment with different masking mechanisms, such as traditional infinity masking, fix-valued masking, and learnable masking.
These variants are evaluated on tasks involving long sequences and imbalanced modality contributions, such as vision-centric perception and knowledge understanding.

\myPara{Response to RQ3: Multimodal Attention Matrix}
we analyze the interaction between duplex attention alignment and modular attention mask by studying their combined effect on the multimodal attention matrix.
This is motivated by the hypothesis that the two components work synergistically to improve multimodal representation learning by enhancing both alignment and attention distribution.
We design experiments that compare the joint use of these components against their individual use, as well as against baseline models without either component.
Tasks such as question answering and multimodal summarization are chosen to simultaneously evaluate alignment and distribution. 
\subsection{Results}
As shown in ~\cref{tab:comp_perception}, ~\cref{tab:comp_cognition}, and ~\cref{tab:comp_emotion}, we demonstrate the main results on 21 popular benchmarks for multimodal perception, cognition, and emotion tasks, respectively.
\begin{table}[!tbp]
\centering
\fontsize{4pt}{4.8pt}\selectfont
\setlength\tabcolsep{2pt}
\renewcommand{\arraystretch}{1.1}
\scalebox{1.49}{
\begin{tabular}{r|p{.314cm}<{\centering}p{.314cm}<{\centering}p{.314cm}<{\centering}p{.314cm}<{\centering}p{.314cm}<{\centering}p{.314cm}<{\centering}p{.314cm}<{\centering}p{.314cm}<{\centering}p{.314cm}<{\centering}}
\multicolumn{1}{c}{{\tiny Model}} &
\multicolumn{9}{c}{{\tiny Cognition}}  \\
Method &
\rotatebox{90}{Avg} &
\rotatebox{90}{Instruction Adherence} &
\rotatebox{90}{Fluency} &
\rotatebox{90}{Coherency} &
\rotatebox{90}{Image-Text relevance} &
\rotatebox{90}{Response Accuracy} &
\rotatebox{90}{Personality Consistency\ } &
\rotatebox{90}{Knowledge Consistency} &
\rotatebox{90}{Tone Consistency} \\
\hline

GPT-4 Turbo          & 1.099 & 1.055 & 1.032 & 1.084 & 1.097 & 1.092 & 1.168 & 1.103 & 1.161 \\
Gemini 1.0 Pro       & 1.021 & 0.999 & 1.007 & 1.028 & 1.009 & 1.013 & 1.052 & 1.013 & 1.050 \\
Claude 3 Opus        & 1.157 & 1.127 & 1.070 & 1.149 & 1.167 & 1.146 & 1.219 & 1.168 & 1.213 \\
QWen-VL-Max          & 1.028 & 1.014 & 1.012 & 1.035 & 1.034 & 1.029 & 1.042 & 1.021 & 1.041 \\

\hline
\rowcolor{gray!10}
\multicolumn{1}{l|}{\textit{Base: Llama-3-Ins-8B}}  &  &  &  &  &  &  &  &  &  \\
Mini-Gemini-HD-8B    & 0.878 & 0.884 & 0.942 & 0.898 & 0.864 & 0.853 & 0.855 & 0.876 & 0.852 \\
LLaVA-NeXT-8B        & 0.968 & 0.971 & 0.988 & 0.980 & 0.966 & 0.967 & 0.966 & 0.964 & 0.939 \\
Cambrian-1-8B        & 0.895 & 0.901 & 0.957 & 0.934 & 0.886 & 0.889 & 0.860 & 0.892 & 0.838 \\
\rowcolor{LightCyan}
\model-8B            &  \textbf{0.972} & \textbf{0.976} & \textbf{0.992} & \textbf{0.985} & \textbf{0.970} & \textbf{0.972} & \textbf{0.970} & \textbf{0.969} & \textbf{0.945} \\

\rowcolor{gray!10}
\multicolumn{1}{l|}{\textit{Cognition-Specialized}}  &  &  &  &  &  &  &  &  &  \\
MMRole-9B            & 0.994 & 0.998 & 1.000 & 0.997 & \textbf{0.993} & 0.987 & 1.000 & \textbf{0.992} & \textbf{0.988} \\
\rowcolor{LightCyan}
\model-8B            &   \textbf{0.995}&\textbf{1.000}&\textbf{1.001}&\textbf{0.999}&\textbf{0.993}&\textbf{0.989}&\textbf{1.001}&{0.991}&\textbf{0.988}      \\

MMRole-9B (In-Test)  & 0.999 & 1.000 & 1.000 & 0.999 & 0.997 & 0.989 & 1.012 & \textbf{0.997} & \textbf{0.997} \\
\rowcolor{LightCyan}
\model-8B (In-Test)  &  \textbf{1.000} & \textbf{1.002} & \textbf{1.001} & \textbf{1.000} & \textbf{0.998} & \textbf{0.992} & \textbf{1.013} & 0.996 & 0.996  \\

MMRole-9B (Out-Test) & 0.981 & 0.992 & 0.999 & 0.993 & 0.979 & 0.981 & 0.963 & 0.977 & 0.962 \\
\rowcolor{LightCyan}
\model-8B (Out-Test) &  \textbf{0.984} & \textbf{0.995} & \textbf{1.002} & \textbf{0.996} & \textbf{0.981} & \textbf{0.983} & \textbf{0.970} & \textbf{0.980} & \textbf{0.965} \\

\end{tabular}
}
\caption{\textbf{Comparison of \model~with other leading MLLMs and cognition task-specialized methods on MMRole benchmark.} The numbers of leading MLLMs come from~\cite{dai2024mmrole}.}
\label{tab:comp_cognition}
\end{table}
\begin{table}[!tbp]
\centering
\fontsize{4pt}{4.8pt}\selectfont
\setlength\tabcolsep{2pt}
\renewcommand{\arraystretch}{1.1}
\scalebox{1.49}{
\begin{tabular}{r |p{.314cm}<{\centering}p{.314cm}<{\centering}p{.314cm}<{\centering}p{.314cm}<{\centering}p{.314cm}<{\centering}p{.314cm}<{\centering}p{.314cm}<{\centering}p{.314cm}<{\centering}p{.314cm}<{\centering}}
\multicolumn{1}{c}{{\tiny Model}} &
\multicolumn{9}{c}{{\tiny Emotion}}  \\
Method &
\rotatebox{90}{Avg} &
\rotatebox{90}{MVSA\textsuperscript{S} (ACC)} &
\rotatebox{90}{MVSA\textsuperscript{S} (F1)} &
\rotatebox{90}{MVSA\textsuperscript{M} (ACC)\qquad\quad\ } &
\rotatebox{90}{MVSA\textsuperscript{M} (F1)} &
\rotatebox{90}{TumEmo (ACC)} &
\rotatebox{90}{TumEmo (F1)} &
\rotatebox{90}{HFM (ACC)} &
\rotatebox{90}{HFM (F1)} \\
\hline

GPT-4V          & 0.633 & 0.507 & 0.570 & 0.609 & 0.631 & 0.608 & 0.612 & 0.764 & 0.765 \\
Gemini 1.0 Pro  & 0.646 & 0.634 & 0.637 & 0.699 & 0.657 & 0.598 & 0.582 & 0.674 & 0.683 \\
Claude 3 Opus   & 0.628 & 0.626 & 0.613 & 0.635 & 0.629 & 0.580 & 0.574 & 0.679 & 0.687 \\
QWen-VL-Max     & 0.643 & 0.647 & 0.645 & 0.669 & 0.627 & 0.565 & 0.595 & 0.696 & 0.701 \\

\hline
\rowcolor{gray!10}
\multicolumn{1}{l|}{\textit{Base: Llama-3-Ins-8B}}  &  &  &  &  &  &  &  &  &  \\
Mini-Gemini-HD-8B   & 0.482 & 0.423 & 0.571 & 0.487 & \textbf{0.643} & 0.246 & 0.395 & 0.498 & \textbf{0.593} \\
LLaVA-NeXT-8B       & 0.576 & 0.591 & 0.593 & 0.617 & 0.607 & 0.547 & 0.533 & 0.572 & 0.551 \\
Cambrian-1-8B       & 0.547 & 0.694 & 0.661 & 0.662 & 0.579 & 0.439 & 0.344 & 0.512 & 0.487 \\
\rowcolor{LightCyan}
\model-8B           & \textbf{0.588} & \textbf{0.702} & \textbf{0.705} & \textbf{0.628} & {0.619} & \textbf{0.559} & \textbf{0.548} & \textbf{0.585} & {0.563} \\

\rowcolor{gray!10}
\multicolumn{1}{l|}{\textit{Emotion-Specialized}}  &  &  &  &  &  &  &  &  &  \\
M\textsuperscript{2}CL & - & 0.755 & 0.742 & 0.732 & 0.705 & 0.688 & 0.687 & - & - \\
MULSER                 & - & 0.757 & 0.755 & 0.739 & 0.738 & 0.775 & 0.775 & - & - \\
CMGCN                  & - & 0.733 & 0.720 & 0.697 & 0.683 & - & - & 0.875 & 0.841 \\
SPFVTE                 & - & 0.806 & 0.801 & 0.799 & 0.788 & - & - & 0.883 & 0.879 \\
\rowcolor{LightCyan}
\model-8B  & \textbf{0.841} & \textbf{0.810} & \textbf{0.803} & \textbf{0.802} & \textbf{0.790} & \textbf{0.778} & \textbf{0.778} & \textbf{0.885} & \textbf{0.881} \\

\end{tabular}
}
\caption{\textbf{Comparison of \model \ with other leading MLLMs as well as emotion task-specialized methods on four emotion benchmarks.}
The reported numbers of emotion-specialized methods come from their official manuscripts.
The missed average performance of emotion-specialized methods due to missed datasets.
}
\label{tab:comp_emotion}
\end{table}

\myPara{Perception Benchmark.}
To assess the effectiveness of our proposed model, we compare it against state-of-the-art Multimodal Large Language Models (MLLMs), including the Mini-Gemini-HD series, the LLaVA-NeXT series, and the Cambrian-1 series.
We conduct a comparison under two settings, where we tune these MLLMs from 8B and 34B scale large foundation models.
Our proposed MODA outperforms other models of similar scale, including LLaVA-NeXT and Cambrian, achieving an average improvement of 1.0 for the base Llama-3-Ins-8B model and 0.9 for the base Hermes2-Yi-34B model.
In vision-centric and OCR tasks, which require fine-grained understanding, MODA consistently performs better, achieving a metric of 66.0 for the vision-centric average and 74.7 for OCR \& Chart tasks.
This highlights the model's ability in tasks demanding fine-grained perception, further reinforcing its superiority.

\myPara{Cognition-specific \& Emotion-specific Benchmark.}
We evaluate a diverse set of MLLMs on both cognition-centric and emotion-centric benchmarks, designed to assess key dimensions of cognitive and emotional understanding across various aspects.
On the cognition benchmark, \model~outperforms open-ended models of Cambrian-1 (0.981) and LLaVA-NeXT (0.979), achieving an average score of 0.996, and performs comparably to close-ended SOTA models like Claude 3 Opus (0.995).
With cognition task-specific tuning, \model~achieves better performance, particularly excelling in fluency (0.999) and personality consistency (0.998).
On emotion benchmarks, \model~outperforms open-ended models like Cambrian-1 (0.628) and LLaVA-NeXT (0.624), with an average score of 0.657, and achieves comparable performance to task-specialized SOTA models like SPFVTE (0.738) and MULSER (0.739).
Notably, \model~excels in TumEmo (0.747) and HFM (0.753) benchmarks, demonstrating strong performance in emotion recognition tasks due to its ability to capture fine-grained emotional features and fine-grained details.
\begin{figure}[!tbp]
  \centering
\includegraphics[width=\linewidth]{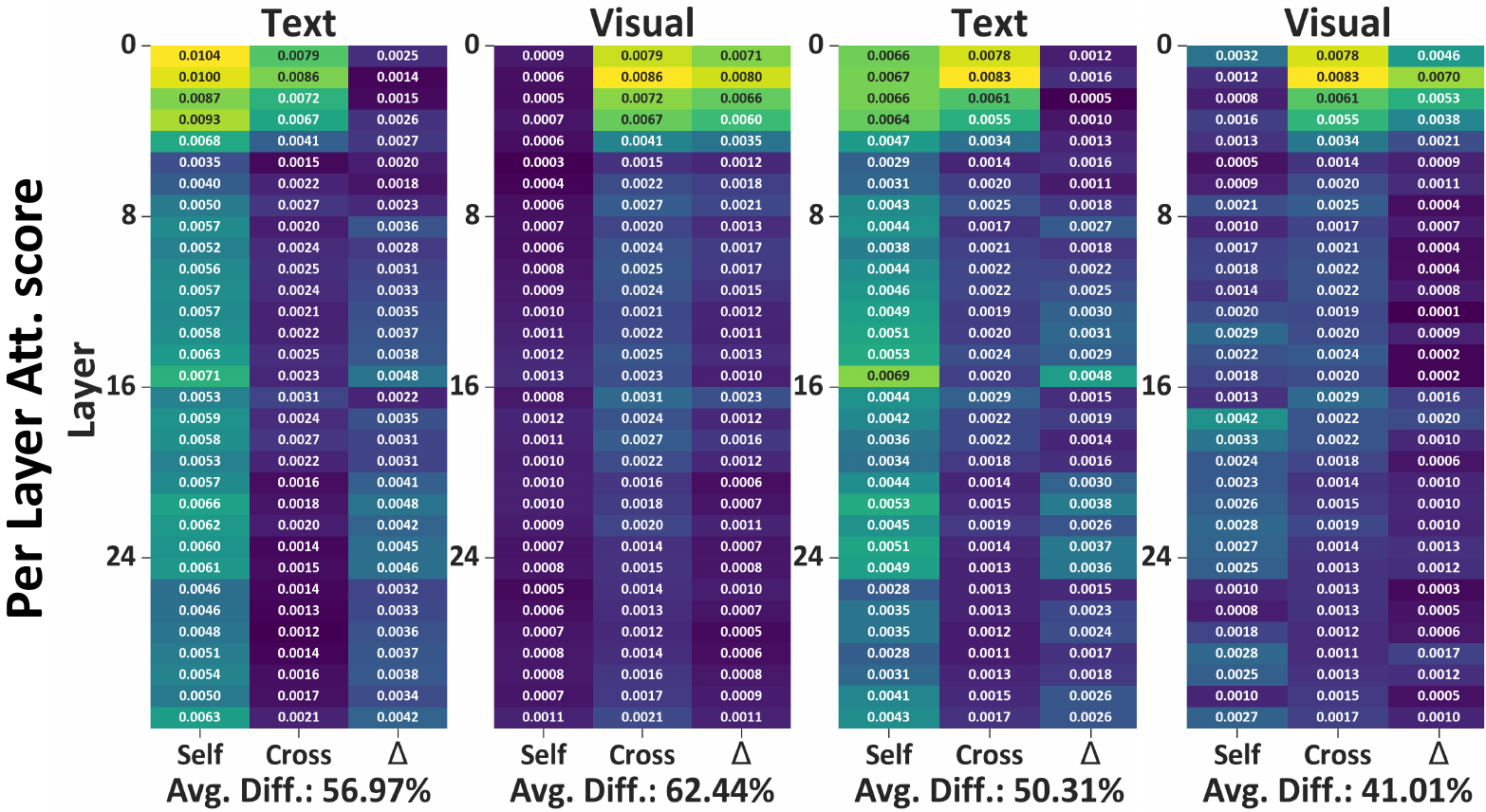}
\put(-209,-8){\small \textbf{(a) Multimodal Attention}}
\put(-72,-8){\small \textbf{(b) MODA}}
  \caption{Analysis of multimodal attention and \model. (a) Multimodal attention: average difference between self and cross-modal attention is 56.97\% for text and 62.44\% for visual. (b) \model: average difference is 50.31\% for text and 41.01\% for visual.}
  \vspace{-5pt}
  \label{fig:MODA_AttentionMap}
\end{figure}
\begin{figure*}[!tbp]
  \centering
  \includegraphics[width=\linewidth]{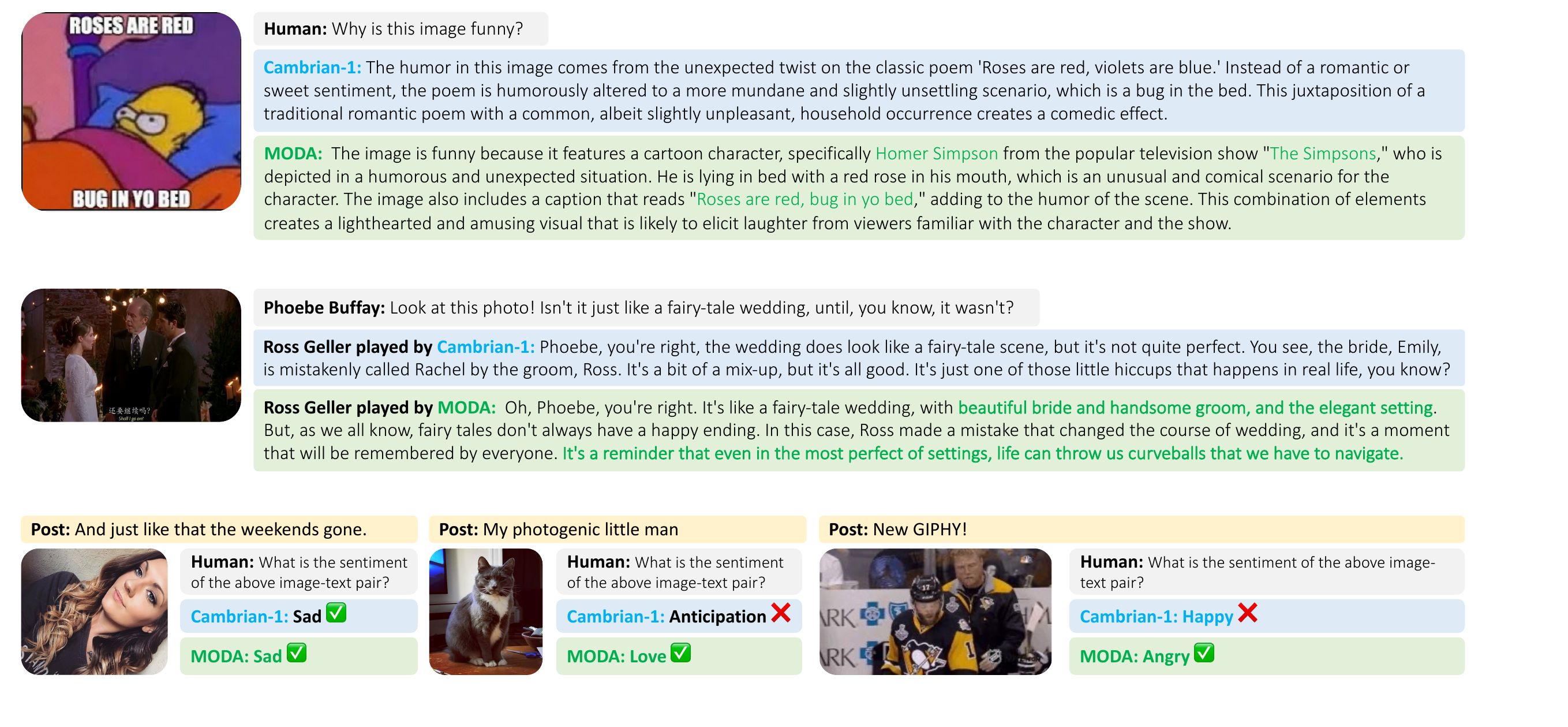}
\put(-287,138){\small \textbf{(a)} Fine-grained Perception}
\put(-303,60){\small \textbf{(b)} Cognitive Conversation Analysis}
\put(-284,-8){\small \textbf{(c)} Emotion Understanding}
  \caption{Visualization results of state-of-the-art (SOTA) Multimodal Large Language Models (MLLMs) and our proposed \model~on fine-grained multimodal tasks, including perception, cognition, and emotion understanding.}
  \label{fig:MODA_Answers}
\end{figure*}

\myPara{Analysis of Attention.}
We analyze the attention distribution to evaluate the effectiveness of the proposed \model~in terms of attention distribution (Fig.~\ref{fig:MODA_AttentionMap}).
We conduct experiments on fine-grained understanding tasks, \textit{i.e.}, emotion recognition based on visual-textual affective cues.
Baseline attention exhibits imbalanced scores, particularly in deeper layers, highlighting its tendency to over-focus on dominant modalities while neglecting others (\textit{i.e.}, visual affective cues).
Our \model achieves consistently high across all layers, preventing degradation and ensuring balanced contributions from all tokens and modalities.
This improvement directly correlates with superior performance on multimodal tasks, particularly those requiring fine-grained understanding, such as TumEmo and OCRBench, demonstrating MODA's ability to avoid attention collapse and capture multimodal interactions effectively.
\begin{figure}[!tbp]
  \vspace{-10pt}
  \centering
  \captionsetup[subfigure]{labelformat=empty}
  \begin{subfigure}[b]{\linewidth}
    \centering
  \includegraphics[width=\linewidth]{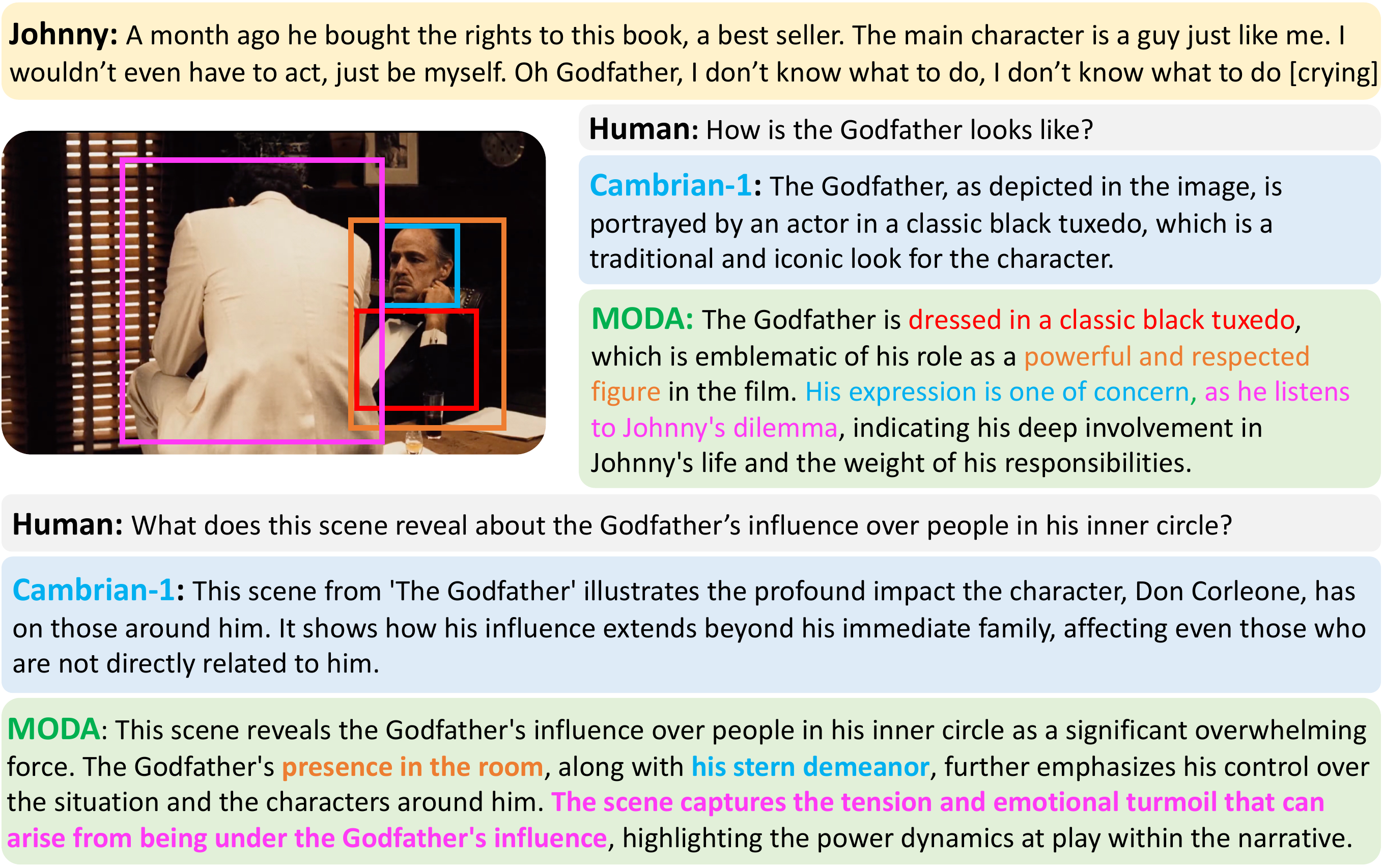}  
  {(a) Understanding \textit{Godfather}}
  \end{subfigure}
  
  \begin{subfigure}[b]{\linewidth}
    \centering
  \includegraphics[width=\linewidth]{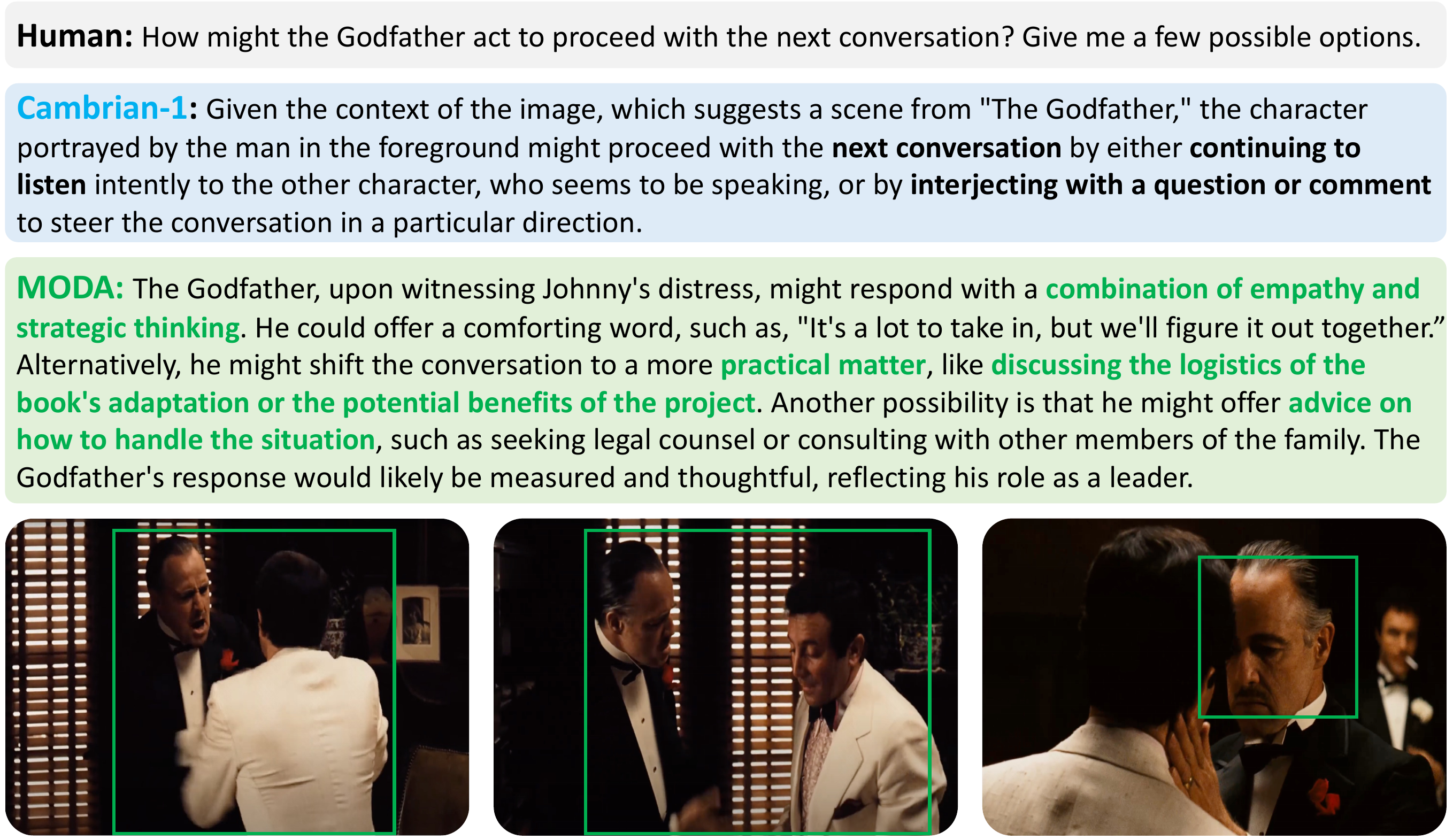}  
  \end{subfigure}
      \vspace{-4pt}
  {(b) Planning for \textit{Godfather}}
    \vspace{-4pt}
  \caption{\model-enabled apps in \textit{The Godfather}. (a) With a deep and fine-grained understanding of conversation, \model~captures both the emotional and cognitive states of the character. (b) \model~further simulates the \textit{Godfather}'s strategic thinking, planning the next steps while considering the character’s traits.}
  \label{fig:discussion}
\end{figure}

\myPara{Visualization}
To highlight the advantages of Modular Duplex Attention (MODA), we design a visualization experiment focusing on output answers, showcasing its capacity to generate fine-grained and accurate responses.
The output answers (Figure~\ref{fig:MODA_Answers}) further demonstrate MODA's superior capacity for fine-grained understanding.
By accurately capturing the humorous twist on the classic poem, MODA has demonstrated its unparalleled ability to reason about fine-grained multimodal details.
These results underscore MODA's transformative potential in advancing the state-of-the-art in multimodal understanding, where it seamlessly integrates information across modalities to achieve exceptional precision in complex cognitive and emotional tasks.

\section{Discussion}
\model-based MLLM can boost various downstream directions.
Here, we envision two potential uses.

\myPara{\textit{Godfather}-centric Understanding.} 
Leveraging the fine-grained understanding of multimodal content,
\model~demonstrates enhanced comprehension capabilities that facilitate human-centric interpretation.
Integrated with the strong sensory grounding capacity, \model~can effectively process and interpret high-level human cognition and emotion.
Notably, we present the first demonstration of a Multimodal Large Language Model (MLLM) that captures emotions expressed by individuals, going beyond simple classification into discrete categories.
As illustrated in Fig.\ref{fig:discussion} (a), \model~successfully captures the micro-expressions of the \textit{Godfather}, reflecting the character’s personality and the cultural context embedded within the scene.

\myPara{Conversation with \textit{Godfather}.}
Further, the comprehensive understanding enabled by \model~enhances its capacity to plan conversations.
We integrate \model~into a conversation system between the human and the agent in three stages: description, analysis, and planning.
As shown in Fig.\ref{fig:discussion} (b), \model~takes the historical conversation context (e.g., video keyframe, transcript, and audio) as input and generates the desired target (e.g., behavior, characteristics, and emotion).
This allows the model to simulate a conversation flow, as in the \textit{Godfather} scenario, where the agent responds with strategic plan based on the established emotional context.

\section{Conclusion}
This work introduces the MOdular Duplex Attention (\model) to tackle attention deficit disorder in multimodal large language models, characterized by inconsistent cross-modal attention and layer decay.
\model~enhances multimodal perception, cognition, and emotion understanding by modularly processing diverse data streams, outperforming existing MLLMs across 21 benchmark datasets.
This advancement not only improves modality alignment but also supports deeper cognitive and emotional insights, with source code and demo available for further exploration.

\section*{Acknowledgement}

This work was supported by the National Natural Science Foundation of China (No. 623B2056), the Natural Science Foundation of Tianjin, China (No.24JCZXJC00040), the Fundamental Research Funds for the Central Universities, the Supercomputing Center of Nankai University (NKSC).
We sincerely thank the reviewer team (cYUZ, XinM, Mf8z, and Ack2) for their invaluable feedback to improve our manuscript.

\section*{Impact Statement}
This paper introduces a novel multimodal attention mechanism designed to enhance Multimodal LLMs for fine-grained content understanding.
However, as with most MLLMs, the quality of \model's output is influenced by the fine-tuning data and the quality of the base models, which may result in the generation of low-quality or hallucinated content.
Such outputs could potentially be harmful, and users are advised to interpret the results with caution, adhering to licensing restrictions, with commercial use explicitly prohibited.
All the personal information is anonymized or obfuscated to ensure confidentiality.

\bibliography{main}
\bibliographystyle{icml2025}

\end{document}